%% file: causal_discovery_using_subsampling.tex
\documentclass{article}
\usepackage{spconf}

\usepackage{amsmath}
\usepackage{graphicx,psfrag,epsf}
\usepackage{enumerate}
\usepackage{url} % not crucial - just used below for the URL 
\usepackage{hyperref}
\usepackage{url} 
\usepackage{amssymb,amsfonts,mathtools}
\usepackage{bm}
\usepackage{lipsum}
\usepackage{color, colortbl}
\usepackage[shortlabels]{enumitem}
\usepackage{etoolbox}
\usepackage{cite}
\usepackage[font=footnotesize]{caption}
\usepackage{cases}
\usepackage{booktabs}
\usepackage{xcolor}
\usepackage{caption}
\usepackage{algorithm}
\usepackage{balance}
\usepackage[noend]{algpseudocode}
\usepackage{subfig}
\usepackage{romannum}
\usepackage{comment}
\usepackage{dsfont}
\usepackage{float}

\makeatletter
\patchcmd{\@makecaption}
  {\scshape}
  {}
  {}
  {}
\makeatletter
\patchcmd{\@makecaption}
  {\\}
  {.\ }
  {}
  {}
\makeatother

\def\P{{ \mathrm{pr} }}

\newcommand{\rom}[1]{\uppercase\expandafter{\romannumeral #1\relax}}

\title{A Subsampling-based Method for Causal Discovery on Discrete Data}
\name{Austin Goddard and Yu Xiang}
\address{  University of Utah\\
50 S Central Campus Dr \#2110 Salt Lake City, UT, USA\\ 
Email: \{austin.goddard, yu.xiang\}@utah.edu}
\input{defns}

\begin{document}
\maketitle
\begin{abstract}
Inferring causal directions on discrete and categorical data is an important yet challenging problem. Even though the additive noise models (ANMs) approach can be adapted to the discrete data, the functional structure assumptions make it not applicable on categorical data. Inspired by the principle that the cause and mechanism are independent, various methods have been developed, leveraging independence tests such as the distance correlation measure. In this work, we take an alternative perspective and propose a subsampling-based method to test the independence between the generating schemes of the cause and that of the mechanism. Our methodology works for both discrete and categorical data and does not imply any functional model on the data, making it a more flexible approach. To demonstrate the efficacy of our methodology, we compare it with existing baselines over various synthetic data and real data experiments. 
\end{abstract}

\section{introduction}

Understanding causal relationships is a fundamental
problem in science. While interventional approaches often lead to accurate identification of causal relationships, they are potentially costly and can even involve invasive procedures. It is thus appealing to perform causal inference based on observational data alone, which has witnessed major advances in recent years~\cite{shimizu2006,Hoyer2009,peters2011causal,zhang2015,peters2016,ghassami2017,Peters2017}, following the pioneering works on structural causal models (SCM) by Pearl~\cite{pearl2000models}. In this work, we focus on inferring causal direction on \emph{discrete and categorical}  data~\cite{peters2011causal,cai2018causal,budhathoki2018accurate,liu2016causal,budhathoki2017mdl,janzing2012information, du2020robust}. Even though the additive noise models (ANMs) can be adapted to the discrete data~\cite{peters2011causal,budhathoki2018accurate}, the additive noise structure may not be suitable for certain applications, especially when the data is categorical in nature. Motivated by the fundamental principle, \emph{independence of the cause and mechanism}~\cite{pearl2000models}, the authors in~\cite{liu2016causal} propose an ingenious approach, which we will refer to as the \emph{DC-causal method} in this work. The main idea is to view the cause $p(x)$ and the mechanism $p(y|x)$ as random variables, and then test the independence of the two via the empirical distance correlation measure~\cite{szekely2007measuring}. Given the flexibility of this approach and its competitive empirical performances on both discrete and categorical data, it has the potential to be adapted to more complex settings. However, we observe that the usage of the empirical distance correlation measure for finite samples can lead to some inherent bias with respect to the support sizes of the cause $X$ and effect $Y$. More importantly, the interpretation of the ``independence of the cause and mechanism" principle is by no means unique, and we propose to test the independence between the \emph{generating scheme} of the cause vs. that of the mechanism. This alternative view allows us to leverage a subsampling method to stabilizes the empirical distance correlation measure. We present various synthetic and real data experiments and show that our method is competitive in comparison with existing baseline methods. 

%turns out to be subtle Inspired by this work, we investigate one variant of it 

%In this work, we focus on the ANMs for discrete data, i.e.,~\cite{peters2011causal} and its variants under different kernel functions.   

%This approach has been extended to handle discrete data~\cite{peters2011causal}, where the authors not only proved the identification results but also proposed an efficient heuristic algorithm. 

%Arguably, the most popular data-driven causal inference framework is the Additive Noise Models (ANMs)~\cite{Hoyer2009,Peters2017,Mooij2016}, where it assumes that the cause and effect are related via a nonlinear function with an additive independent noise. Intuitively, the nonlinearity breaks the symmetry between the two potential cause-effect directions, making causal discovery possible. 

\section{The DC-causal method: Two Caveats}

To set the stage, let us consider a pair of \emph{discrete} random variables $(X,Y)\sim p(x, y)$ (that is, $\P((X,Y)=(x,y)) = p(x,y)$), where the support sizes of $X$ and $Y$ are $|\Xc|$ and $|\Yc|$, respectively. For simplicity of presentation, we write $\xv = (x_1,..., x_{|\Xc|})$ and $\yv = (y_1,..., y_{|\Yc|})$. We use capital letters $(X, Y)$ (or $(\Xv, \Yv)$) for random variables (or vectors), while the lowercase ones $(x, y)$ (or $(\xv, \yv)$) for the corresponding realizations. Also we write $p(\xv): = (p(x_1),...,p(x_{|\Xc|}))$ and $p(\yv): = (p(y_1),...,p(y_{|\Yc|}))$.
Let us focus on the causal direction from $x$ to $y$, (i.e., $x\to y$) to describe the DC-causal method. The main idea is to view $(p(x_i), p(\yv|x_i))$, $i\in\{1,..., |\Xc|\}$, as one sample of a random variable pair $(p(X), p(\yv|X))$, where $X\sim p(x)$ and $p(\yv|X):= (p(y_1|X),..., p(y_{|\Yc|}|X))$. Hence there are $|\Xc|$ samples in total. 

We now discuss the implications, focusing on explaining the effect of viewing each realization as one sample. First note that $p(X)$ is one-dimensional and $p(\yv|X)$ has dimension $|\Yc|$. The distributions of these two random variables can be written as follows: for $i\in\{1,..., |\Xc|\}$, 
\begin{align*}
	\P\bigl(p(X)=p(x_i)\bigr) &= p(x_i),\\
	\P\bigl(p(\yv|X) = p(\yv|x_i)\bigr) & = p(x_i).
\end{align*}
They can never be independent of each other since both of them depend on $X$. So in order to remove this common dependency on $X$, the authors in \cite{liu2016causal} implicitly apply randomized transformations to these two random variables so that they are \emph{uniformly} distributed over the same set of realizations, and this is accomplished by \emph{forcing each realization to be one sample}. As a consequence, it is now sufficient to sample each realization once rather than according to weights $p(x_i)$. More concretely, consider the following two randomized transformations $U_x(\cdot)$ and $U_{y|x}(\cdot)$ such that 
 \begin{align*}
	\P\bigl(U_x(p(X))=p(x_i)\bigr) &= 1/|\Xc|,\\
	\P\bigl(U_{y|x}(p(\yv|X)) = p(\yv|x_i)\bigr) & = 1/|\Xc|.
\end{align*}
Based on this, \emph{the main postulate in~\cite{liu2016causal} is that $x$ causes $y$ is equivalent to $U_x(p(X))$ and $U_{y|x}(p(\yv|X)$ being independent of each other}. The problem of testing $x\to y$ thus boils down to testing the independence by inputting $|\Xc|$ samples $(p(x_i), p(\yv|x_i))$, $i\in\{1,..., |\Xc|\}$, to the empirical distance correlation measure for the independence test~\cite{szekely2007measuring}. Similarly, to test the reverse direction $y\to x$, one uses $|\Yc|$ samples $(p(y_j), p(\xv|y_j))$, $j\in\{1,..., |\Yc|\}$. Now we argue that there are two caveats of the DC-causal method~\cite{liu2016causal}.

  % \subsection{Support-size Bias}
  \smallskip
   \noindent\underline{\bf Support-size bias.} First, we start with a \emph{support-size bias} associated with the method. It turns out that for finite sample size, the distance correlation estimate has an inherent bias. Roughly speaking, for a pair of independent variables, the distance correlation measure decreases (towards $0$) as the support sizes grow. \emph{As a result, the DC-causal method tends to claim the one with the smaller (or larger) support size as the effect (or cause).} Furthermore, it has been observed that this phenomena exists in other methods intended for discrete and categorical variables where the ANM assumption is relaxed. (see Section VII in~\cite{budhathoki2017mdl} for detailed discussions about support bias in the method we later refer to as CISC). Note that an unbiased version of the distance correlation measure is available~\cite{szekely2013distance}, however, it is not stable when the sample size is small.

 %  \subsection{Inputs of the Distance Correlation Measure.}
    \smallskip
      \noindent\underline{\bf Inputs of the Distance Correlation Measure.} A more critical caveat, in our view, is the  inputs of the distance correlation measure. As we have explained above, the DC-causal method aims to test the independence between $U_x(p(X))$ and $U_{y|x}(p(\yv|X)$ for the $x\to y$ direction and thus treat each realization $(p(x_i), p(\yv|x_i))$ as one sample, leading to a total of $|\Xc|$ samples. Similarly, the reverse direction is tested based on $|\Yc|$ samples. However, for discrete random variables, either $|\Xc|$ or $|\Yc|$ or both can be too small to obtain reliable estimates. More importantly, we interpret the \emph{independence of the cause and mechanism} principle~\cite{pearl2000models} as that the cause $p(\xv)$ and the mechanism $p(\yv|\xv)$ are generated independently so that changing one does not affect the other. Therefore, it seems more natural to us to consider testing the independence between the underlying generating schemes of the cause and the mechanism. 
       
       Specifically, consider two unknown random generating schemes $\Gv_{\xv}$ and $\Gv_{\yv|\xv}$ that generate $p(\xv)$ and $p(\yv|\xv)$, respectively, where $p(\xv)$ is viewed as one realization of the cause and $p(\yv|\xv)$ the mechanism. Given a data set $\Dc=\{(x_{i},y_{i})\}_{i=1}^{N}$, we therefore hope to test the independence between $\Gv_{\xv}$ and $\Gv_{\yv|\xv}$ by inputting different samples of $p(\xv)$ and $p(\yv|\xv)$, achieved via subsampling, into the empirical distance correlation measure. We discuss this in more detail in the next section.

       \section{Subsampling with distance correlation}
       Suppose we have $m$ random samples $(\Xv_i, \Yv_i)$, $1\le i\le m$, where $\Xv_i=(X_{i,1},..., X_{i,p})$ and $\Yv_i=(Y_{i,1},..., Y_{i,q})$ coming from the joint distribution of $(\Xv,\Yv)$. Define $a_{i,j}=\|\Xv_i - \Xv_j \|$ for  $i,j=1,..., m$ and
\begin{align*}
	a_{i,\cdot}=\frac{1}{m}\sum_{j=1}^m a_{i,j}, \quad a_{\cdot,j}=\frac{1}{m}\sum_{j=1}^m a_{,j},\quad a_{\cdot,\cdot}=\frac{1}{m}\sum_{i,j=1}^m a_{i,j}.
\end{align*}
Then let $A_{i,j}=a_{i,j} - a_{i,\cdot} - a_{\cdot,j} + a_{\cdot,\cdot}$ and similarly construct $A_{i,j}$ based on $b_{i,j}=\|\Yv_i - \Yv_j \|$. The empirical distance covariance  is defined as 
$\Cc_m(\Xv,\Yv) = \frac{1}{m^2}\sum_{i,j=1}^m A_{i,j}B_{i,j}$
The distance correlation $\Rc_m(\Xv,\Yv)$ is the just the normalized $\Cc_m(\Xv,\Yv)$, and we refer the readers to~\cite{szekely2007measuring} for more details.

  \begin{algorithm}[H]
    %\footnotesize
%        \captionsetup{font=footnotesize}
	    \caption{Subsampling with distance correlation \label{alg:1}
}
	\begin{algorithmic}
    	\State {\textbf{Input:} Data set $\Dc=\{(x_{i},y_{i})\}_{i=1}^{N}$, subsampling probability $q$, number of subsampled datasets $m$}
    	\State {\textbf{Output:} 
   $x \to y$ or $y \to x$, 
    	}
\Procedure{test}{$x \to y$}
\State {1. Subsample $m$ datasets $\hat{\Dc}_1$, ... $\hat{\Dc}_m$ from $\Dc$, where $\hat{\Dc}_i$ includes each $(x_{i},y_{i})$ independently with probability $q$.}
 \State {2. For each dataset $i$, compute the corresponding empirical distribution $(\hat{p}_i(\xv), \hat{p}_i(\yv|\xv))$}. 
 \State {3. Treat $\{(\hat{p}_i(\xv), \hat{p}_i(\yv|\xv))\}_{i=1}^m$ as $m$ random samples of $(p(\xv), p(\yv|\xv))$ and compute the empirical distance correlation $\Rc_m(\Gv_{\xv}, \Gv_{\yv|\xv})$}.
             \EndProcedure
\Procedure{test}{$y \to x$}
\EndProcedure
\If {$\Rc_m(\Gv_{\xv}, \Gv_{\yv|\xv}) < \Rc_m(\Gv_{\yv}, \Gv_{\xv|\yv})$}
\State Output: $x \to y$ 
\Else
 \State Output: $y \to x$
\EndIf
	\end{algorithmic}
\end{algorithm}

As we have alluded to in the previous section, we propose a \emph{subsampling-based DC method} presented in Algorithm~1. The main idea is to first bootstrap (via subsampling) an ensemble of $m$ empirical estimates of $p(\xv)$ and $p(\yv|\xv)$, and then use the distance correlation measure to test the independence using these $m$ samples. However, it should be noted that a support-size bias still remains but in a different manner and we try to provide some insights on this as follows. Suppose that $|\Xc|<|\Yc|$ and given a fixed dataset, then the $m$ empirical estimates of $p(x)$ will be more similar to each other than that of $p(y)$, making the former closer to be ``uniform". As a consequence, it is more likely that $\Rc_m(\Gv_{\xv}, \Gv_{\yv|\xv}) < \Rc_m(\Gv_{\yv}, \Gv_{\xv|\yv})$, creating a bias to make $y$ more likely to be the effect. \emph{Therefore, it is reasonable to only test cases when $|\Xc|=|\Yc|$, which is what we will do in our experiment section.}

We now discuss the choices of $m$ and $p$ based on our empirical experiments. We choose $p$ according to the following rule: $p^* =\min \{p_f, p_b\}$, 
where $p_f$ denotes the $p$ that minimizes the empirical distance correlation calculated based on the forward channel, i.e.,  $p_f = \arg\min_p \Rc_m(\Gv_{\xv}, \Gv_{\yv|\xv})$. Similarly, $p_b$ is defined for the backward model. One can also optimize $p$ under different criteria, such as maximizing the gap between $\Rc_m(\Gv_{\xv}, \Gv_{\yv|\xv})$ and $\Rc_m(\Gv_{\yv}, \Gv_{\xv|\yv})$. We choose to adopt the former one due to its empirical performance. Note that our  subsampling approach produces a random output, due to the randomness from the subsampling procedure. We thus choose large enough $m$ so that the results become stable, and we illustrate this in our synthetic data experiments in the next section.

\section{experiments}
\begin{figure}[h!]
    \subfloat{%
      \includegraphics[width=0.24\textwidth]{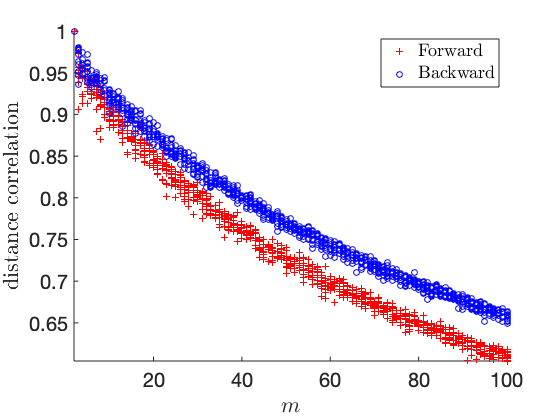}
    }
    \subfloat{%
      \includegraphics[width=0.24\textwidth]{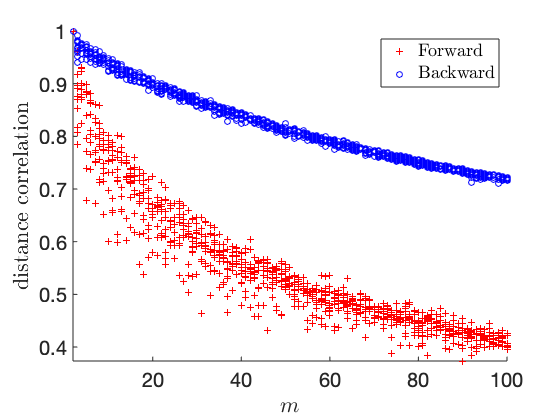}
    }
    \caption{Distance correlation vs. $m$ in Experiment~1 (left) and~2 (right).}
    \label{fig:calc_m}
\end{figure}

In our experiments, we compare our method (referred to as SUB in the plots) with the DC-causal method as well as various existing baseline methods: DR~\cite{peters2011causal}, CISC~\cite{budhathoki2017mdl}, ICGI~\cite{janzing2012information}, HCR~\cite{cai2018causal}, and ACID~\cite{budhathoki2018accurate}. The ground truth is $x\to y$ for all the cause-effect pairs in our experiments.  For the synthetic data, we choose $m=100$ and it is justified in Fig.~\ref{fig:calc_m}. For each real data experiments, we choose the $m$ by plotting the distance correlations for both forward and backward directions, and then choose large enough $m$ so that the two directions do not overlap. $p$ is chosen based on the criteria mentioned above. $p_f$ and $p_b$ are determined by selecting the smallest of $10$ linear spaced $p$'s between $0.01$ and $0.99$.

\subsection{Synthetic Data Experiments}
In this section, we revisit two synthetic data experiments from~\cite{liu2016causal}. As we shall illustrate below, the original setups favor the forward direction ($x\to y$) by \emph{making $p(x)$ (or $p(y|x)$) closer to a uniform distribution}, since $\Rc_n(p(x), p(y|x))$ is close to $0$ if $p(x)$ or $p(y|x)$ is close to uniform. In light of this, we propose to modify the setups to make the two directions less distinguishable. In all synthetic data experiments, accuracy results are averaged over $1000$ separate datasets. Each dataset contains $2000$ samples.

\smallskip
\noindent{\bf Experiment 1: } First, we summarize the setup of Section~5.1 in~\cite{liu2016causal}. The cause and effect are related via an additive noise model: $Y = f(X) + N$, where the noise $N$ is independent of $X$, and $\Xc=\{1,2,...,30\}$. (I) $p(x)$ is obtained by randomly generating a vector (of length $|\Xc|$) with each entry being an integer between  $[1,|\Xc|/4]$ and then normalizing it. (II) $f$ is a random mapping from $\Xc$ to $\Yc_0=\{1,2,...,30\}$.

\begin{figure}[h!]
\begin{center}
    \subfloat{
      \includegraphics[width=0.2\textwidth]{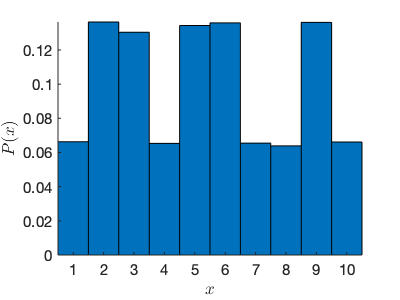}
    }
    \subfloat{
      \includegraphics[width=0.2\textwidth]{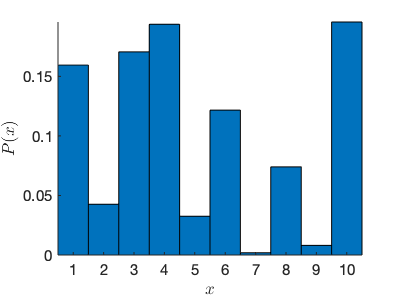}
    }
    \caption{$p(x)$ from the modified setup (left) vs. from the original setup (right).}
    \label{fig:5p1_compare_dists}
    \end{center}
        \vspace{-1em}
\end{figure}

Now we make some observations. In (I), sampling $p(x)$ from a discrete distribution on a smaller interval tends to produce many repeating probabilities in $p(x)$ (see Fig.~\ref{fig:5p1_compare_dists} for a simple illustration). Note that probabilities close to constant in $p(x)$ result in a smaller empirical distance correlation in the forward direction than the backward. Because of (II), multiple $x$'s
     will be mapped to the same $y$, resulting in repeated conditional distributions in
     $p(y|x)$ and thus a more uniform $p(y|x)$. 
     
     We propose to make the following changes.
(a) Given the support sizes $|\Xc|$ and $|\Nc|$, we generate $p(x)$ and $p(n)$ from the (continuous) uniform distribution over $[0,1]$. 
(b) The function $f$ is constructed using a random but one-to-one mapping from $|\Xc|$ to $|\Yc_0|$. As a minor change for the sake of making $|\Xc|=|\Yc|$ easily, we adopt $Y = (f(X) + N )\text{ mod } |\Yc_0|$, inspired by the cyclic models in~\cite{peters2011causal}. Note that the curves are not sensitive to this change.   

\begin{figure}[h!]
     \centering
     \includegraphics[width=0.33\textwidth]{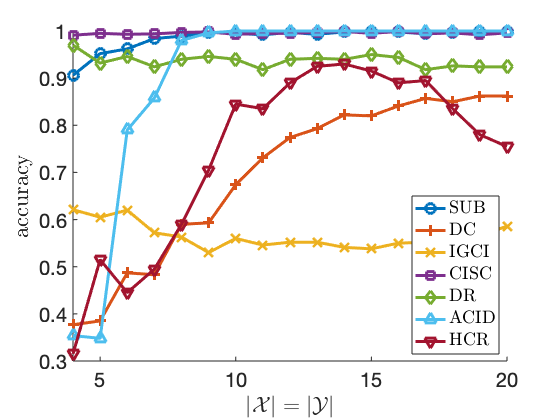}
     \caption{Experiment 1}
     \label{fig:exp5p1}
\end{figure}

Results of the modified setup are presented in Fig. \ref{fig:exp5p1} and we fix the support of $|\Nc|$ to be  $\{-2,-1,0,1,2\}$. Our subsampling method achieves near perfect accuracy even when the support sizes are small. DR consistently achieves around $90\%$ accuracy. Except on support sizes less than $8$, ACID achieves close to $100\%$ accuracy. IGCI performs only slightly better than a random coin flip. Like the subsampling method, the CISC method also achieves near optimal results. However, the subsampling method's reliability can be seen by examining the relative gap (defined below) between each method's forward and backward scores, that is, the empirical distance correlation for the subsampling method and the stochastic complexity for CISC. Here we define the relative gap as $|s_f - s_b|/\max\{s_f,s_b\}$, where $s_f$ and $s_b$ denote the forward score and backward score, respectively. The subsampling method has a relative gap of $0.13$ while the CISC method has $2.5\times 10^{-4}$. Since it has been observed that the CISC method also has an inherent bias and favors the variable with the smaller support size as the cause, we assert that the subsampling method holds an edge over the CISC method. 

\smallskip
\noindent{\bf Experiment 2: } We start with summarizing the setup of Section~5.2 in~\cite{liu2016causal}. Both $p(x)$ and $p(y|x)$ are generated randomly, and $p(x)$ is obtained in the same way as in Experiment~1, while $p(y|x)$ is selected using a reference set, which contains $|\Xc|/4$ distributions. Each distribution in the reference set is generated in the same manner as that of $X$ with the exception that $|\Yc|$ samples are taken. Then, for $x \in X$, $p(y|x)$ is selected randomly from the reference set. 

Following the same reasoning as in the previous experiment, we propose the following changes. (a) We generate $p(x)$ and $p(y|x)$ from the (continuous) uniform distribution over $[0,1]$. (b) The reference set is removed. Each conditional distribution $p(y|x)$ is created in the same way as $p(x)$.

\begin{figure}[h!]
     \centering
     \includegraphics[width=0.33\textwidth]{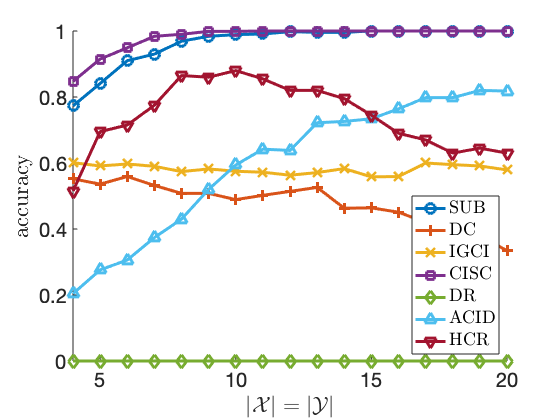}
     \caption{Experiment 2}
     \label{fig:exp5p2}
      %   \vspace{-1em}
\end{figure}

The results are presented in Fig. \ref{fig:exp5p2}. As in Experiment $1$, the subsampling method and CISC are highly accurate at inferring the correct causal direction. Though, due to the relative gap, we still assert the subsampling method as superior. The results of DR and ACID are less accurate compared to the results in Experiment $1$.  We attribute this to the fact that the datasets in Experiment 2 do not emit an ANM-like structure. DC and IGCI perform only slightly better than a coin flip. 

         \vspace{-1em}

\subsection{Real Data Experiments}
We test our method on real world datasets that are available at https://webdav.tuebingen.mpg.de/cause-effect/. Given that most of them are continuous variables, we need to preprocess them by rounding in order to reduce their support sizes. Specifically, \emph{we investigate the variables with at most two digits after the decimal point, and simply round them to the nearest integers}. After the preprocessing step, we focus on the cases where the cause and effect have the same or similar support sizes that are less than $50$. There are $8$ such datasets in total and we investigate all of them individually as follows.

\smallskip
\noindent{\bf Longitude/latitude vs. temperature.} These are the 3rd and 20th datasets, and they were collected in $349$ locations in Germany from $1961$ to $1990$. The three variables \emph{longitude, latitude, and temperature} have support sizes $10$, $9$ and $11$, respectively. Our subsampling method infers the correct directions in both cases, and ACID identifies the correct direction for the latitude vs. temperature dataset. All the other method fail.   

\smallskip
\noindent{\bf Population growth and food consumption.}  This is the 76th dataset, and it is taken from food security statistics provided by Food and Agriculture Organization of the United Nations during $1990$ to $2000$. $x$ denotes the average annual rate of change of population, while $y$ the average annual rate of change of total dietary consumption for total population. After preprocessing, they have support sizes $13$ and $16$, respectively. Four methods are able to identify the correct direction: our method, ACID, IGCI, and CISC. 
       %  \vspace{-1em}

\smallskip
\noindent{\bf Some continuous-valued datasets.} There are a few datasets (as listed below) that the support sizes of cause and effect are close to each other after some preprocessing steps. However, the raw datasets have many digits after the decimal points. Thus we prefer to treat them as continuous rather than discrete, and we discuss the cases as follows. The financial datasets (65th, 66th, and 67th) have more than $10$ digits after the decimal point. Thus even with rounding (or $\textsf{round}(100\times a)$ for a real-valued number $a$ as in~\cite{liu2016causal}), we consider them  to be continuous in nature and thus the methods for discrete data might not be appropriate. In fact, we observe that the inferred causal direction changes with different resolutions $k$, which can be defined through $\textsf{round}(10^k\times a)$. We therefore consider the inferred direction as unreliable. Two similar datasets are: \emph{the logarithm of employment and logarithm of population} (the 84th dataset) and \emph{the logarithm of the performance measure and characteristics of CPUs} (the 100th dataset).

\bibliographystyle{IEEEtran}
\balance

%\newpage
\bibliography{ref}
	
                 \end{document}

%% file: defns.tex
\usepackage{xspace}
\usepackage{bbm}
\usepackage{mathrsfs}
\newcommand{\Cc}{\mathcal{C}}

\newcommand{\Dc}{\mathcal{D}}

\newcommand{\Nc}{\mathcal{N}}

\newcommand{\Rc}{\mathcal{R}}

\newcommand{\Xc}{\mathcal{X}}
\newcommand{\Yc}{\mathcal{Y}}

\newcommand{\Gv}{{\bf G}}
\newcommand{\Xv}{{\bf X}}
\newcommand{\Yv}{{\bf Y}}

\newcommand{\xv}{{\bf x}}
\newcommand{\yv}{{\bf y}}

\let\P\relax
\DeclareMathOperator\P{\sf P}
\def\textiid{i.i.d.\@\xspace}
\newcommand\iid{\ifmmode\text{ i.i.d. } \else \textiid \fi}

%% file: causal_discovery_using_subsampling.bbl
\begin{thebibliography}{10}
\providecommand{\url}[1]{#1}
\csname url@samestyle\endcsname
\providecommand{\newblock}{\relax}
\providecommand{\bibinfo}[2]{#2}
\providecommand{\BIBentrySTDinterwordspacing}{\spaceskip=0pt\relax}
\providecommand{\BIBentryALTinterwordstretchfactor}{4}
\providecommand{\BIBentryALTinterwordspacing}{\spaceskip=\fontdimen2\font plus
\BIBentryALTinterwordstretchfactor\fontdimen3\font minus
  \fontdimen4\font\relax}
\providecommand{\BIBforeignlanguage}[2]{{%
\expandafter\ifx\csname l@#1\endcsname\relax
\typeout{** WARNING: IEEEtran.bst: No hyphenation pattern has been}%
\typeout{** loaded for the language `#1'. Using the pattern for}%
\typeout{** the default language instead.}%
\else
\language=\csname l@#1\endcsname
\fi
#2}}
\providecommand{\BIBdecl}{\relax}
\BIBdecl

\bibitem{shimizu2006}
S.~Shimizu, P.~O. Hoyer, A.~Hyv{\"a}rinen, and A.~Kerminen, ``A linear
  non-gaussian acyclic model for causal discovery,'' \emph{Journal of Machine
  Learning Research}, vol.~7, no. Oct, pp. 2003--2030, 2006.

\bibitem{Hoyer2009}
P.~O. Hoyer, D.~Janzing, J.~M. Mooij, J.~Peters, and B.~Sch{\"o}lkopf,
  ``Nonlinear causal discovery with additive noise models,'' in \emph{Advances
  in neural information processing systems}, 2009, pp. 689--696.

\bibitem{peters2011causal}
J.~Peters, D.~Janzing, and B.~Scholkopf, ``Causal inference on discrete data
  using additive noise models,'' \emph{IEEE Transactions on Pattern Analysis
  and Machine Intelligence}, vol.~33, no.~12, pp. 2436--2450, 2011.

\bibitem{zhang2015}
K.~Zhang, B.~Huang, J.~Zhang, B.~Sch{\"o}lkopf, and C.~Glymour, ``Discovery and
  visualization of nonstationary causal models,'' \emph{arXiv preprint
  arXiv:1509.08056}, 2015.

\bibitem{peters2016}
J.~Peters, P.~B{\"u}hlmann, and N.~Meinshausen, ``Causal inference by using
  invariant prediction: identification and confidence intervals,''
  \emph{Journal of the Royal Statistical Society: Series B (Statistical
  Methodology)}, vol.~78, no.~5, pp. 947--1012, 2016.

\bibitem{ghassami2017}
A.~Ghassami, S.~Salehkaleybar, N.~Kiyavash, and K.~Zhang, ``Learning causal
  structures using regression invariance,'' in \emph{Advances in Neural
  Information Processing Systems}, 2017, pp. 3011--3021.

\bibitem{Peters2017}
J.~Peters, D.~Janzing, and B.~Sch\"olkopre, \emph{Elements of Causal Inference:
  Foundations and Learning Algorithms}.\hskip 1em plus 0.5em minus 0.4em\relax
  Cambridge, MA, USA: MIT Press, 2017.

\bibitem{pearl2000models}
J.~Pearl, ``Models, reasoning and inference,'' \emph{Cambridge, UK: Cambridge
  University Press}, 2000.

\bibitem{cai2018causal}
R.~Cai, J.~Qiao, K.~Zhang, Z.~Zhang, and Z.~Hao, ``Causal discovery from
  discrete data using hidden compact representation,'' in \emph{Advances in
  neural information processing systems}, 2018, pp. 2666--2674.

\bibitem{budhathoki2018accurate}
K.~Budhathoki and J.~Vreeken, ``Accurate causal inference on discrete data,''
  in \emph{2018 IEEE International Conference on Data Mining (ICDM)}.\hskip 1em
  plus 0.5em minus 0.4em\relax IEEE, 2018, pp. 881--886.

\bibitem{liu2016causal}
F.~Liu and L.~Chan, ``Causal inference on discrete data via estimating distance
  correlations,'' \emph{Neural computation}, vol.~28, no.~5, pp. 801--814,
  2016.

\bibitem{budhathoki2017mdl}
K.~Budhathoki and J.~Vreeken, ``{MD}l for causal inference on discrete data,''
  in \emph{2017 IEEE International Conference on Data Mining (ICDM)}.\hskip 1em
  plus 0.5em minus 0.4em\relax IEEE, 2017, pp. 751--756.

\bibitem{janzing2012information}
D.~Janzing, J.~Mooij, K.~Zhang, J.~Lemeire, J.~Zscheischler, P.~Daniusis,
  B.~Steudel, and B.~Sch{\"o}lkopf, ``Information-geometric approach to
  inferring causal directions,'' \emph{Artificial Intelligence}, vol. 182, pp.
  1--31, 2012.

\bibitem{du2020robust}
K.~Du, A.~Goddard, and Y.~Xiang, ``On the robustness of causal discovery with
  additive noise models on discrete data,'' in \emph{2020 Data Compression
  Conference}.\hskip 1em plus 0.5em minus 0.4em\relax IEEE, 2020, pp. 365--365.

\bibitem{szekely2007measuring}
G.~J. Sz{\'e}kely, M.~L. Rizzo, N.~K. Bakirov \emph{et~al.}, ``Measuring and
  testing dependence by correlation of distances,'' \emph{The annals of
  statistics}, vol.~35, no.~6, pp. 2769--2794, 2007.

\bibitem{szekely2013distance}
G.~J. Sz{\'e}kely and M.~L. Rizzo, ``The distance correlation t-test of
  independence in high dimension,'' \emph{Journal of Multivariate Analysis},
  vol. 117, pp. 193--213, 2013.

\end{thebibliography}
